\documentclass{article}
\usepackage{iclr2027_conference,times}
\usepackage[T1]{fontenc}
\usepackage{amsmath,amssymb,amsthm,booktabs,array,tabularx,graphicx}
\usepackage[table]{xcolor}
\usepackage{algorithm,algpseudocode}
\usepackage{hyperref,url}
\hypersetup{hidelinks}
\usepackage{amsmath}
\usepackage{bbm}
\newcolumntype{Y}{>{\raggedright\arraybackslash}X}

\title{What Should SFT Learn? Conditional Information Gain for Generalizable Post-Training}

\author{
    Yuxuan Gu\thanks{Equal contribution. Code is available at \url{https://github.com/NiuKequan/CIG-SFT-public}.}
    \quad
    Kequan Niu\footnotemark[1]
    \quad
    Xiaocheng Feng
    \quad
    Yun Li
    \quad
    Yiqiao Yang
    \quad
    Bing Qin
    \\
    Harbin Institute of Technology \\
    \texttt{\{yxgu,keniu,xcfeng\}@ir.hit.edu.cn}
}
\iclrfinalcopy

\begin{document}
\maketitle

\fancyhead[L]{Preprint}

\begin{abstract}
Supervised fine-tuning (SFT) remains a widely used approach for post-training large language models, because fixed demonstrations can directly introduce desired reasoning strategies and behavioral patterns. Yet forcing a model to imitate externally provided trajectories may also interfere with behaviors already supported by pretraining, limiting generalization beyond the training distribution. Recent dynamic SFT methods mitigate this issue by reweighting token-level supervision according to model-side signals such as predictive probability, prior support, or distributional compatibility. However, \emph{model compatibility is not equivalent to learning value}: a token that is weakly supported by the current model may either be an undesirable mismatch or precisely the task-specific decision that the demonstration is intended to teach.
We propose \textbf{CIG-SFT}, an information-theoretic framework that estimates the task-conditioned information carried by each demonstrated token. Under teacher forcing, we compare the token likelihood from a frozen reference model with and without access to the task input while keeping the gold solution prefix fixed. The resulting conditional information gain quantifies how much the task itself reduces uncertainty about the demonstrated decision. We then convert this information into a KL-regularized projection of the current policy and derive an equivalent scalar-weighted SFT objective whose logit gradient exactly matches that of the projected target.
Across mathematical reasoning and code-generation benchmarks, CIG-SFT consistently outperforms standard SFT and recent adaptive fine-tuning baselines, with particularly strong improvements in generalization to challenging evaluation distributions, while requiring neither additional training data nor online rollouts.
\end{abstract}

\section{Introduction}
\label{sec:introduction}

Large language models (LLMs) acquire broad and increasingly general capabilities during large-scale pretraining, ranging from language understanding and generation to reasoning and problem solving \citep{brown2020language,chowdhery2023palm}. Since base models are not inherently optimized to express capabilities in the forms required by diverse downstream tasks, post-training plays a critical role in transforming a general-purpose base model into one that exhibits desired task- and application-specific behaviors \citep{chung2024scaling}. By providing explicit behavioral supervision or optimization signals, post-training can improve instruction following \citep{wei2021finetuned,ouyang2022training}, shape how models reason and make intermediate decisions \citep{lightman2024let,guo2025deepseek}, and encourage adherence to task-specific response structures and output conventions \citep{chu2025sft}. These adaptations are particularly important for reasoning-intensive applications, where success depends not only on whether a model possesses the underlying capability, but also on whether it can invoke and organize that capability into an effective solution process. Accordingly, supervised fine-tuning (SFT) and reinforcement learning (RL) have become two central paradigms for modern LLM post-training, providing complementary mechanisms for shaping model behavior beyond pretraining \citep{ouyang2022training,schulman2017proximal,guo2025deepseek}.

However, such adaptation introduces a fundamental tension between \textit{acquiring target behaviors} and \textit{preserving the capabilities that LLMs already support}. 
It is especially evident in post-training methods that learn from fixed, externally provided data, including SFT, offline preference optimization \citep{rafailov2023direct}, and offline reinforcement learning \citep{snell2022offline}. Their key advantage is behavioral coverage: by directly optimizing on curated demonstrations or feedback, these methods can expose the model to reasoning steps and behavioral patterns that it may rarely produce on its own. 
Yet this same property can also create a mismatch between externally specified supervision and the model's existing predictive behavior, biasing LLM behavior toward the demonstrated solution patterns at the expense of catastrophic forgetting of capabilities inherited from pretraining \citep{kotha2024understanding,li-etal-2024-revisiting}.
On-policy approaches address the same tension from the opposite direction,
which generate training trajectories from the evolving policy itself and thus make the supervision naturally better aligned with LLMs' predictive behavior distribution, such as reinforcement learning \citep{schulman2017proximal,ouyang2022training,guo2025deepseek}, self-taught reasoning \citep{zelikman2022star}, on-policy distillation \citep{agarwal2024policy}, and on-policy self-distillation \citep{zhao2026self}.
Nevertheless, this compatibility comes at the cost of behavioral coverage, as useful strategies encoded in external demonstrations may be missed when the current policy rarely visits them \citep{agarwal2024policy,zhao2026self}.

Recent work mitigates this trade-off by retaining fixed demonstrations while selectively modulating their supervision at the token level. DFT reweights each target according to its probability under the current model \citep{wu2025generalization}, where subsequent methods stabilize this reweighting through distributional anchoring \citep{zhu2025anchored}, estimate token support from a frozen pretrained model \citep{wang2026prift}, or explicitly characterize whether a target is compatible with the model-induced distribution \citep{zhang2026idft}. Despite different formulations, they share a common principle: \emph{the strength of supervision is largely determined by how well a demonstrated token is supported by the model's predictive distribution}.
However, predictive compatibility is not equivalent to learning value. A weakly supported target may indicate an undesirable distributional mismatch, but it may also correspond to precisely the task-specific decision that the demonstration is intended to teach. Conversely, a highly supported target may reflect a useful model-consistent behavior, or simply a trivial continuation already determined by the preceding gold trajectory. Model support alone cannot distinguish what it \emph{should acquire} from what it \emph{should avoid}, or what has been sufficiently captured. 


Therefore, we quantify the learning value of each demonstrated token from an information-theoretic perspective, viewing information gain as the reduction in uncertainty induced by an additional observation
\citep{shannon1948mathematical,lindley1956measure,cover2006elements}.
In supervised fine-tuning, the task input specifies the behavior to be learned, while the preceding response tokens also contribute to predicting the next token.
Model-support signals computed from the full context may conflate task-induced information with continuation-induced predictability.
We disentangle these two sources by measuring the additional information contributed by the task input beyond the preceding context.
In detail, we propose \textbf{CIG-SFT}, which translates pointwise conditional information gain into principled supervised updates.
First, we estimate the task-conditioned information carried by each demonstrated token via comparing its log-likelihood ratio under the frozen base model with and without the task input while keeping the preceding response fixed.
The resulting ratio is a pointwise conditional information quantity, which we normalize by task-free surprisal to obtain a bounded information-gain ratio.
Next, we treat this value as a token-level reward and derive a KL-regularized projection of the current policy, producing an information-dependent target distribution that balances information acquisition against policy deviation.
We show that the gradient induced by this projected target can be reproduced exactly by a scalar-weighted SFT objective.
In this way, CIG-SFT provides a principled path from task-conditioned information to controlled policy adaptation, while retaining fixed demonstrations and requiring neither additional training data nor online rollouts.

We evaluate CIG-SFT on mathematical reasoning and code generation across multiple model scales.
CIG-SFT consistently improves over standard SFT and strong adaptive fine-tuning baselines, while comparing favorably with reinforcement-learning methods without requiring additional data or online rollouts.
Its gains persist across domains, training-set sizes, and hyperparameter settings, with efficiency in some low-data regimes.
Ablation and training-dynamics analyses further show that CIG-SFT benefits from the separation between frozen information attribution and current-policy-aware proximal adaptation, producing more selective
supervision and smaller parameter updates.
Overall, our results demonstrate that task-conditioned information provides a useful complement to predictive compatibility for generalizable supervised post-training.
Our contributions are threefold:
(\textbf{i}) we identify task-conditioned information as a complementary signal to full-context model support for selective SFT;
(\textbf{ii}) we derive an information-proximal target and show that its optimization admits an exactly gradient-equivalent token-weighted objective; and
(\textbf{iii}) we demonstrate consistent gains across model scales and domains, with data efficiency and more selective parameter updates.


\section{Related Work}
\label{sec:related-work}


\paragraph{Supervised fine-tuning and adaptive objectives.}
Supervised fine-tuning (SFT) remains a standard component of LLM post-training, using dense token-level supervision on fixed demonstrations to adapt pretrained models to desired behaviors \citep{ouyang2022training,chung2024scaling}. Despite its effectiveness, directly fitting demonstrations can lead to overfitting, memorization, or degradation of capabilities outside the fine-tuning distribution \citep{kotha2024understanding,li-etal-2024-revisiting,chu2025sft}. Recent work revisits the uniform weighting of token-level supervision in standard SFT.
Dynamic Fine-Tuning (DFT) rescales each target-token loss using its probability under the current model \citep{wu2025generalization}. Anchored SFT (ASFT) introduces KL-based anchoring to control distributional drift \citep{zhu2025anchored}. PriFT derives token weights from a frozen pretrained model to decouple prior support from the evolving optimization trajectory \citep{wang2026prift}, while IDFT uses distributional statistics to identify supervision better aligned with the model-induced distribution \citep{zhang2026idft}. Related approaches such as ProFit and InfoSFT further explore probability- or information-based token selection and weighting \citep{liu2026profit,sabbaghi2026infosft}. Collectively, these methods show that non-uniform supervision can improve the generalization of fixed-demonstration SFT.
Our method, CIG-SFT, is closest to this line of work, but differs in the signal used to define token-level learning value. Existing adaptive SFT methods primarily derive their weights from predictive confidence, prior support, or distributional typicality under the complete context. CIG-SFT instead uses conditional information gain to quantify how much the task input contributes to each target beyond what the preceding context already reveals.

\paragraph{Reinforcement learning and on-policy post-training.}
Reinforcement learning provides a major paradigm for LLM post-training, where training trajectories are sampled from the evolving policy rather than fixed in advance. PPO-based RLHF has been widely used to optimize language models from preference-derived rewards \citep{schulman2017proximal,ouyang2022training}, while GRPO and subsequent variants enable scalable outcome-based optimization for reasoning tasks \citep{shao2024deepseekmath,guo2025deepseek,yu2026dapo}. Because supervision applies to states the current policy actually visits, on-policy optimization naturally matches the model's generative distribution and avoids forcing the model to imitate externally specified trajectories.
However, the same policy dependence also limits behavioral coverage. Because on-policy supervision comes from trajectories generated by the current policy, behaviors the model rarely discovers are unlikely to receive strong learning signals. This matters especially when external demonstrations encode reasoning strategies or behavioral patterns not yet well represented in the current policy. Empirical studies of reasoning RL provide concrete manifestations of this limitation, including the absence of informative successful rollouts on difficult prompts and the progressive loss of exploration diversity during training \citep{cui2025entropy,yu2026dapo,wang-etal-2026-targeted}. On-policy methods improve the model's compatibility, but do not guarantee exposure to the behaviors that post-training aims to introduce.

\section{Method}
\label{sec:method}

In this section, we present how to derive and implement CIG-SFT, an information-theoretic framework for selectively learning from fixed demonstrations.
Our central principle is to separate two questions that are conflated in conventional token-weighted SFT:
\emph{(i) how much task-specific information does a demonstrated behavior contain, and
(ii) given this information, how far should the current policy move toward that behavior?}
CIG-SFT answers the first question through pointwise conditional information gain under a frozen reference model, and the second through a KL-regularized projection of the current policy.
We further show that the resulting projected update can be realized exactly, at the logit-gradient level, by a scalar-weighted SFT objective.

\begin{figure}[t]
    \centering
    \includegraphics[width=1\linewidth]{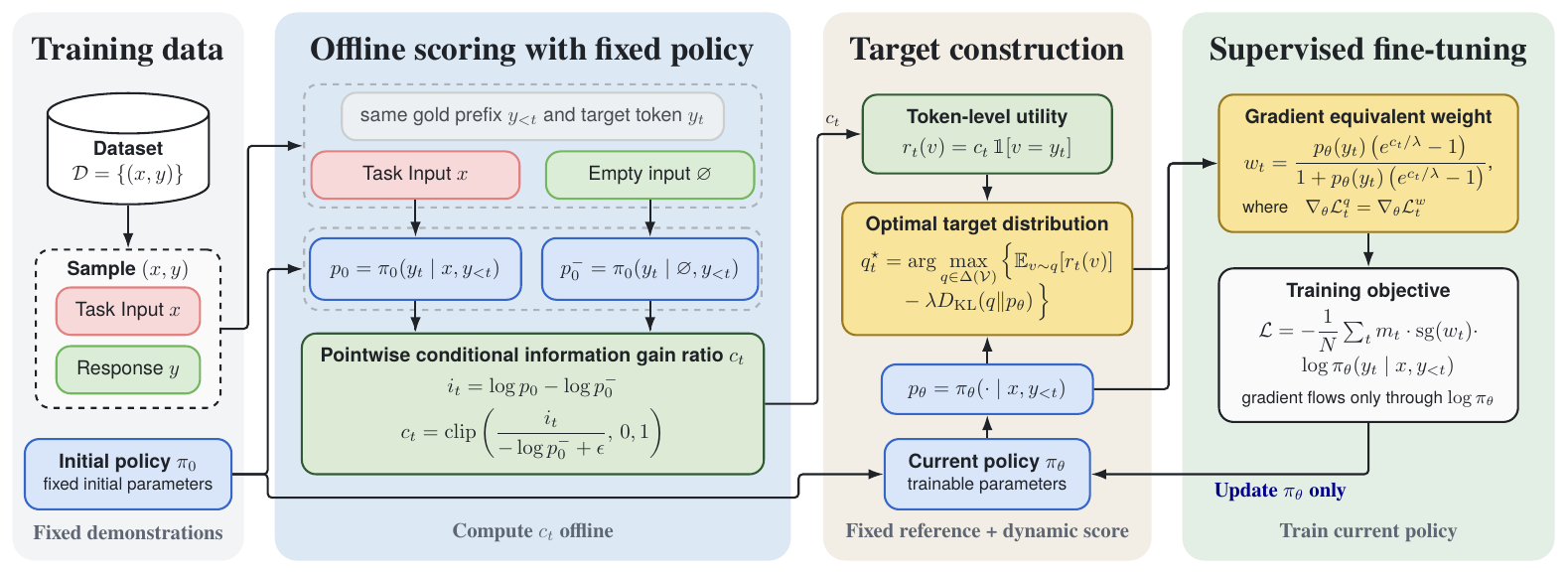}
    \caption{\textbf{Overview of CIG-SFT.}
A frozen initial policy estimates token-level conditional information gain from fixed demonstrations. The information score and current policy jointly define an information-proximal target distribution, which is converted into a gradient-equivalent token weight for supervised optimization. During training, we update only the current policy.}
    \label{fig:questft-overview}
\end{figure}

\subsection{Preliminaries}
\label{sec:method-preliminaries}

Let $\mathcal{D}=\{(x,y)\}$ denote a dataset of fixed demonstrations, where $x$ is the task input and
$y=(y_1,\ldots,y_T)$ is the response with $T$ tokens.
The preceding context for the $t$-th target token is denoted as $y_{<t}$.
We use $\pi_0$ to denote the frozen reference policy and $\pi_\theta$ the trainable policy initialized from the same pretrained model.
The objective of standard supervised fine-tuning is to minimize:
\begin{equation}
\mathcal{L}_{\mathrm{SFT}}
=
-\frac{1}{N}
\sum\nolimits_{t}
m_t
\log
\pi_\theta(y_t\mid x,y_{<t}),
\label{eq:sft}
\end{equation}
where $m_t\in\{0,1\}$ masks tokens participating in training and
$N=\sum_t m_t$.
Dynamic or selective SFT methods generalize
eq.~(\ref{eq:sft}) by assigning different learning strengths to individual target tokens:
\begin{equation}
\mathcal{L}_{\mathrm{dyn}}
=
-\frac{1}{N}
\sum\nolimits_t
m_t\,
\operatorname{sg}[\alpha_t]
\log
\pi_\theta(y_t\mid x,y_{<t})
+
\beta\mathcal{R}(\theta),
\label{eq:dynamic-sft}
\end{equation}
where $\operatorname{sg}[\cdot]$ denotes stop-gradient, $\mathcal{R}(\theta)$ optionally regularizes the updated policy with strength $\beta$, and
$\alpha_t$ controls token-level supervision that is instantiated differently across methods:
\begin{equation}
\alpha_t=\left\{
\begin{aligned}
&\pi_\theta(y_t\mid x,y_{<t})\\[3pt] 
&u(
\pi_0(\cdot \mid x,y_{<t}),y_t)\\[3pt]
&\pi_\theta(y_t\mid x,y_{<t})^{\gamma_t}\\[3pt]
&\pi_\theta(y_t\mid x,y_{<t})
\end{aligned}\right.
\qquad\qquad
\mathcal{R}(\theta)=\left\{
\begin{aligned}
&0\\[3pt] 
&0\\[3pt]
&0\\[3pt]
&D_{\mathrm{KL}}
\left(
\pi_{0}
\|
\pi_\theta
\right)
\end{aligned}\right.
\qquad\qquad\quad
\begin{aligned}
&\text{DFT}\\[3pt] 
&\text{PriFT}\\[3pt]
&\text{IDFT}\\[3pt]
&\text{ASFT}
\end{aligned}
\end{equation}
where $u(\cdot)$ is a prior-support function computed from the frozen LLM, while $\gamma_t$ is a
distribution-compatibility coefficient determined by the target-token probability and the predictive uncertainty of the current policy.
Despite their different instantiations, these methods share a common principle: token-level supervision is primarily determined by full-context model support, which does not disentangle the contribution of the task input $x$ from that of the preceding context $y_{<t}$.

\subsection{Pointwise Conditional Information Gain}
\label{sec:pcig}

To disentangle the influence of task evidence, we compare the same demonstrated token under task-conditioned and task-free contexts.
Using the frozen reference policy $\pi_0$, we define
\begin{equation}
p_0
=
\pi_0(y_t\mid x,y_{<t}),
\qquad
p_0^{-}
=
\pi_0(y_t\mid \varnothing,y_{<t}).
\label{eq:two-contexts}
\end{equation}
where $\varnothing$ denotes an empty task input under the same input format.
The only difference between $p_0$ and $p_0^{-}$ is the
availability of task information $x$.
This contrast is motivated by conditional mutual information \citep{cover2006elements}.
For random variables $X$, $Y_t$, and $Y_{<t}$,
\begin{equation}
I(X;Y_t\mid Y_{<t})
=
\mathbb{E}
\left[
\log
\frac{
p(Y_t\mid X,Y_{<t})
}{
p(Y_t\mid Y_{<t})
}
\right].
\label{eq:cmi}
\end{equation}
Accordingly, inspired by the quantity inside the expectation of eq.~(\ref{eq:cmi}), that is, the pointwise \emph{conditional information gain}, we define our information gain $i_t$ for each demonstrated token as
\begin{equation}
i_t
=
\log p_0
-
\log p_0^{-}
=
\log
\frac{
\pi_0(y_t\mid x,y_{<t})
}{
\pi_0(y_t\mid\varnothing,y_{<t})
}.
\label{eq:cig}
\end{equation}
Since the exact marginal $\pi_0(y_t | y_{<t})$ is intractable, we use the
task-free prediction $\pi_0(y_t\mid \varnothing, y_{<t})$ as an approximation \citep{Xu2020A,pmlr-v162-ethayarajh22a}.
In information theory, the \emph{surprisal} of an event with probability $p$ is $-\log p$, which quantifies how unexpected the event is \citep{shannon1948mathematical}. Therefore, $i_t$ measures the reduction in target-token surprisal from observing $x$ beyond what the preceding context already explains. We normalize this reduction by the task-free surprisal as
%
\begin{equation}
c_t
=
\operatorname{clip}
\left(
\frac{
i_t
}{
-\log p_0^{-}+\epsilon
},
0,1
\right),
\label{eq:cigr}
\end{equation}
where $\epsilon>0$ is a small constant for numerical stability when the task-free surprisal approaches zero.
$c_t$ quantifies the ratio of the target token's task-free
surprisal explained by the task input: $c_t=0$ indicates no positive surprisal reduction, while $c_t$ approaching $1$ indicates that the task input explains nearly all of the surprisal remaining after conditioning on the preceding context.
Since $c_t$ depends only on the frozen reference policy, it can be computed once and cached before fine-tuning.

\subsection{Information-Proximal Projection}
\label{sec:information-projection}

Conditional information gain quantifies task-side evidence for a demonstrated decision, but does not specify how far the current policy should move toward that decision.
Motivated by reward-weighted policy optimization \citep{peters2007reinforcement,wu2026generalization,zhu2025anchored}, we interpret the normalized information score $c_t$ as a token-level utility for the demonstrated decision:
\begin{equation}
r_t(v)=c_t\,\mathbbm{1}[v=y_t], \quad y_t\in \mathcal{V},\quad\forall v \in \mathcal{V},
\end{equation}
where $\mathcal{V}$ denotes the LLM vocabulary.
Instead of directly maximizing this utility, we constrain the update to remain close to the current policy, following regularized policy optimization \citep{peters2010relative,pmlr-v37-schulman15}. 
Let $p_\theta = \pi_\theta(\cdot\mid x,y_{<t})$,
we obtain the projected target distribution as
\begin{equation}
q_t^\star
=
\arg\max_{q\in\Delta(\mathcal{V})}
\left\{
\mathbb{E}_{v\sim q}[r_t(v)]
-
\lambda
D_{\mathrm{KL}}
\left(
q
\|
p_\theta
\right)
\right\},
\label{eq:projection}
\end{equation}
where $\Delta(\mathcal{V})=\left\{q:\mathcal{V}\rightarrow[0,1]\;\middle|\;\sum_{v\in\mathcal{V}}q(v)=1\right\}$ denotes the probability simplex over the vocabulary
and $\lambda>0$ controls the strength of the proximity constraint.
The optimization in eq.~(\ref{eq:projection}) admits a closed-form solution through exponential tilting of the current policy \citep{peters2010relative},
\begin{equation}
q_t^\star(v)
=
\frac{
p_\theta(v)
\exp\!\left(r_t(v)/\lambda\right)
}{
Z_t
}.
\label{eq:exp-tilting}
\end{equation}
Since $r_t(v)$ is nonzero only for the demonstrated token $y_t$, the normalization constant simplifies to
\begin{equation}
\begin{aligned}
Z_t
&=
\sum\nolimits_{v\in\mathcal{V}}
p_\theta(v)
\exp\!\left(r_t(v)/\lambda\right)=p_\theta(y_t)e^{c_t/\lambda} + \sum\nolimits_{v\in\mathcal{V}\setminus y_t}
p_\theta(v)\\
&=p_\theta(y_t)e^{c_t/\lambda}+1-p_\theta(y_t)=1+p_\theta(y_t)\left(e^{c_t/\lambda}-1\right),
\end{aligned}
\end{equation}
and the projected probabilities assigned to the target $y_t$ and non-target token $v\neq y_t$ become
\begin{equation}
q_t^\star(y_t)
=
\frac{
p_\theta(y_t) e^{c_t/\lambda}
}{
1+p_\theta(y_t)\left(e^{c_t/\lambda}-1\right)
}, \quad\text{and}\;\;
q_t^\star(v)
=
\frac{
p_\theta(v)
}{
1+p_\theta(y_t)\left(e^{c_t/\lambda}-1\right)
}.
\label{eq:qstar-target}
\end{equation}
The projection adjusts the current policy according to the task-conditioned information score.
When $c_t=0$, $q_t^\star$ reduces exactly to the current policy, while increasing $c_t$ or decreasing $\lambda$ produces a larger shift toward the demonstrated decision that remaining proximal to the current policy.

\subsection{Gradient-Equivalent Objective Realization}
\label{sec:information-consistent}


The projection yields the optimal target distribution $q_t^\star$ under eq.~(\ref{eq:projection}), which naturally serves as a soft supervision target for updating the trainable policy. We minimize the cross-entropy loss:
%
\begin{equation}
\mathcal{L}_t^{q}
=
-\sum\nolimits_{v\in\mathcal{V}}
\operatorname{sg}\!\left[q_t^\star(v)\right]
\log p_\theta(v).
\label{eq:q-target-loss}
\end{equation}
%
Taking the gradient with respect to $\theta$, we have
\begin{equation}
\nabla_\theta\mathcal{L}_t^q
=
-\sum\nolimits_{v\in\mathcal V}
q_t^\star(v)
\nabla_\theta\log p_\theta(v).
\end{equation}
Since $p_\theta(v)$ is a normalized discrete distribution, there is
\begin{equation}
\sum\nolimits_{v\in\mathcal V}
p_\theta(v)\nabla_\theta\log p_\theta(v)
=
\sum\nolimits_{v\in\mathcal V}\nabla_\theta p_\theta(v)
=0,
\end{equation}
and hence
\begin{equation}
\sum\nolimits_{v\in\mathcal{V}\setminus y_t}
p_\theta(v)\nabla_\theta\log p_\theta(v)
=
-p_\theta(y_t)\nabla_\theta\log p_\theta(y_t).
\label{eq:123}
\end{equation}
Using equations~(\ref{eq:qstar-target}) and (\ref{eq:123}), we obtain
\begin{equation}
\nabla_\theta \mathcal{L}_t^q
=
-\left[
\frac{
p_\theta(y_t)(e^{c_t/\lambda}-1)
}{
1+p_\theta(y_t)(e^{c_t/\lambda}-1)
}
\right]
\nabla_\theta\log p_\theta(y_t).
\label{eq:q-gradient}
\end{equation}

This gradient is exactly reproduced by the weighted hard-target objective 
\begin{equation}
\mathcal{L}_t^w
=
-\operatorname{sg}(w_t)\log p_\theta(y_t),\ \text{as} \;\;\nabla_\theta\mathcal{L}_t^{q}
=
\nabla_\theta\mathcal{L}_t^{w},
\label{eq:weighted-token-loss}
\end{equation}
where the token-wise update coefficient $w_t$ is
\begin{equation}
w_t
=
\frac{
p_\theta(y_t)\left(e^{c_t/\lambda}-1\right)
}{
1+p_\theta(y_t)\left(e^{c_t/\lambda}-1\right)
}.
\label{eq:information-weight}
\end{equation}
It is worth noting that equations~(\ref{eq:q-target-loss}) and
(\ref{eq:weighted-token-loss}) are not generally equal in value.
They are \emph{gradient-equivalent} with respect to $\theta$ when their targets or weights are detached.
Aggregating over all supervised tokens, we get the final training objective with equivalent gradients as
\begin{equation}
\mathcal{L}_{\mathrm{CIG\text{-}SFT}}
=
-\frac{1}{N}
\sum\nolimits_t
m_t\,
\operatorname{sg}(w_t)
\log
\pi_\theta(y_t\mid x,y_{<t}),
\quad \text{where }\, N=\sum\nolimits_t m_t.
\label{eq:final-objective}
\end{equation}

\section{Experiments}
\label{sec:experiments}

\subsection{Experimental Setup}
\label{sec:exp-setup}

\paragraph{Models and datasets.}
We evaluate CIG-SFT across mathematical reasoning and code generation.
For mathematical reasoning, we use Qwen2.5-Math-1.5B and Qwen2.5-Math-7B \citep{yang2024qwen2}, trained on a deduplicated 30K subset of NuminaMath-CoT \citep{li2024numinamath}.
We evaluate on MATH500 \citep{lightman2024let},
Minerva Math \citep{lewkowycz2022solving},
OlympiadBench \citep{he2024olympiadbench},
AIME 2024, AIME 2025, and AMC 2023\footnote{\url{https://maa.org/maa-invitational-competitions/}}.
For code generation, we use Qwen2.5-Coder-3B
\citep{hui2024qwen2}, trained on a 10K subset of the code training pool, and evaluate on HumanEval+ and MBPP+
\citep{liu2023your}, LiveCodeBench v5/v6
\citep{jain2025livecodebench}, BigCodeBench-Full
\citep{zhuo2025bigcodebench}, and DS-1000
\citep{lai2023ds}.




\paragraph{Baselines.}
We compare with representative supervised fine-tuning, offline preference optimization, and on-policy reinforcement learning methods.
For supervised fine-tuning, we include standard SFT and several recent adaptive objectives:
Dynamic fine-tuning (DFT) \citep{wu2025generalization}, which reweights target tokens according to their probabilities under the current policy;
In-distribution fine-tuning (IDFT) \citep{zhang2026idft}, which combines target-token probability with predictive uncertainty to adapt the token-level learning signal;
Anchored Supervised Fine-Tuning (ASFT) \citep{zhu2025anchored}, which augments probability-based reweighting with distributional anchoring;
and PriFT \citep{wang2026prift}, for which we evaluate both PriFT-prob and PriFT-mass using token statistics from a frozen pretrained reference model.
We additionally compare against offline preference optimization methods, including Direct Preference Optimization (DPO) \citep{rafailov2023direct} and SimPO \citep{meng2024simpo}, 
as well as on-policy reinforcement learning methods,
including Group Relative Policy Optimization (GRPO)
\citep{shao2024deepseekmath} and DAPO \citep{yu2026dapo}.

\paragraph{Training and evaluation settings.}
For mathematical reasoning, all supervised fine-tuning methods are trained for one epoch using AdamW \citep{loshchilov2017decoupled}, cosine learning-rate decay, a warm-up ratio of $0.1$, a learning rate of $5\times10^{-5}$, a global batch size of $128$, and a maximum sequence length of $2048$.
For CIG-SFT, we set $\lambda=0.3$, which controls the strength of the conditional-information-gain signal when constructing the projected token-level target distribution. We keep this value fixed across all experiments unless otherwise specified.
For evaluation, we sample $16$ responses per problem at temperature $1.0$ with a maximum generation length of $4096$.
We report Avg@16, the average correctness percentage across the $16$ sampled responses, and Pass@16, the fraction of problems for which at least one of the $16$ responses is correct.


\subsection{Main Results}
\label{sec:exp-main}

\paragraph{Mathematical reasoning.}
Tables~\ref{tab:math7} and~\ref{tab:math15} summarize the mathematical reasoning results.
CIG-SFT consistently achieves the strongest aggregate performance across both model scales. 
On Qwen2.5-Math-1.5B, CIG-SFT reaches \textbf{$31.5\%$ Avg@16} and \textbf{$58.6\%$ Pass@16}, improving over standard SFT by $17.7/16.3$ and over DFT by $5.2/9.6$, respectively.
It also outperforms the strongest prior supervised baseline, PriFT-prob, by 1.2 points in Avg@16 and 2.4 points in Pass@16.
The same trend becomes more pronounced on Qwen2.5-Math-7B, where CIG-SFT achieves \textbf{38.5\% Avg@16} and \textbf{63.5\% Pass@16}, exceeding PriFT-mass by $2.1/0.3$ points and DFT by $6.8/12.4$ points.
The improvements are distributed across benchmarks of different difficulty, with particularly clear gains on MATH500, OlympiadBench, and AIME.


\begin{table}
\centering
\setlength{\tabcolsep}{3.5pt}
\caption{Mathematical reasoning results with Qwen2.5-Math-7B. We report Avg@16 / Pass@16.}
\label{tab:math7}
\begin{tabular}{l|c|cccccc}
\toprule
\midrule
\textbf{Method} & \textbf{Average} & \textbf{MATH500} & \textbf{Minerva} & \textbf{Olympiad} & \textbf{AIME24} & \textbf{AIME25} & \textbf{AMC23} \\
\midrule
Base & $17.2/55.6$ & $39.4/88.2$ & $11.7/46.7$ & $14.8/57.2$ & $\phantom{0}8.5/46.7$ & $\phantom{0}2.9/20.0$ & $25.8/75.0$ \\
\midrule
\midrule
SFT & $19.7/49.4$ & $53.0/87.2$ & $16.9/48.9$ & $19.2/55.1$ & $\phantom{0}2.3/13.3$ & $\phantom{0}1.7/16.7$ & $25.3/75.0$ \\
DFT & $31.7/51.1$ & $68.9/86.0$ & $23.9/43.4$ & $33.5/56.3$ & $11.7/23.3$ & $\phantom{0}5.8/20.0$ & $46.6/77.5$ \\
IDFT & $30.1/49.9$ & $69.3/85.0$ & $25.7/42.6$ & $34.1/53.5$ & $\phantom{0}6.7/23.3$ & $\phantom{0}3.1/20.0$ & $41.7/87.5$\\
ASFT & $30.3/60.1$ & $67.8/\underline{92.2}$ & $27.2/\underline{53.7}$ & $33.7/\underline{64.6}$ & $\phantom{0}8.1/33.3$ & $\phantom{0}3.1/26.7$ & $41.7/90.0$ \\
PriFT-prob & $35.8/62.5$ & $72.3/\underline{92.2}$ & $\underline{29.6}/\textbf{54.8}$ & $36.8/\textbf{65.3}$ & $14.4/43.3$ & $\phantom{0}\underline{7.5}/26.7$ & $53.9/\underline{92.5}$ \\
PriFT-mass & $\underline{36.4}/\underline{63.2}$ & $\underline{74.0}/91.8$ & $27.8/52.9$ & $\underline{36.9}/61.8$ & $\underline{15.0}/46.7$ & $\phantom{0}\underline{7.5}/\textbf{33.3}$ & $\textbf{57.3}/\textbf{92.5}$ \\
\rowcolor{cyan!40!green!10!white} CIG-SFT\rule{0pt}{2.ex} & $\textbf{38.5}/\textbf{63.5}$ & $\textbf{75.2}/\textbf{92.4}$ & $\textbf{30.3}/53.3$ & $\textbf{39.4}/64.3$ & $\textbf{19.8}/\textbf{53.5}$ & $\textbf{10.6}/\underline{30.0}$ & $\underline{55.5}/87.5$ \\
\midrule
\bottomrule
\end{tabular}
\end{table}

\begin{table}
\centering
\setlength{\tabcolsep}{3.5pt}
\caption{Mathematical reasoning results with Qwen2.5-Math-1.5B. We report Avg@16 / Pass@16.}
\label{tab:math15}
\begin{tabular}{l|c|cccccc}
\toprule
\midrule
\textbf{Method} & \textbf{Average} & \textbf{MATH500} & \textbf{Minerva} & \textbf{Olympiad} & \textbf{AIME24} & \textbf{AIME25} & \textbf{AMC23} \\
\midrule
Base & $16.5/51.6$ & $39.0/86.0$ & $10.2/41.2$ & $19.2/52.1$ & $\phantom{0}4.8/26.7$ & $1.9/23.3$ & $23.6/80.0$ \\
\midrule
\midrule
\multicolumn{8}{l}{\bfseries\itshape Off-policy Reinforcement Learning}\\[2pt]
DPO & $17.3/49.7$ & $40.0/86.0$ & $11.1/44.1$ & $19.7/55.0$ & $\phantom{0}4.2/33.3$ & $0.8/10.0$ & $28.0/80.0$ \\
SimPO & $17.0/50.9$ & $41.2/86.0$ & $10.9/43.8$ & $19.6/52.4$ & $\phantom{0}4.2/26.7$ & $1.7/16.7$ & $24.2/80.0$ \\
\midrule
\midrule
\multicolumn{8}{l}{\bfseries\itshape On-policy Reinforcement Learning}\\[2pt]
GRPO & $28.0/56.2$ & $63.2/90.4$ & $22.7/\underline{51.1}$ & $29.1/\textbf{62.1}$ & $\phantom{0}7.5/30.0$ & $4.8/23.3$ & $40.9/80.0$ \\
DAPO & $\underline{30.5}/\underline{57.7}$ & $\underline{67.2}/\textbf{91.6}$ & $\textbf{24.9}/\textbf{52.5}$ & $30.2/\underline{60.7}$ & $\phantom{0}9.6/23.3$ & $\textbf{5.8}/\textbf{33.3}$ & $45.0/85.0$ \\
\midrule
\midrule
\multicolumn{8}{l}{\bfseries\itshape Supervised Fine-Tuning}\\[2pt]
SFT & $13.8/42.3$ & $41.5/81.8$ & $12.1/40.4$ & $12.0/45.6$ & $\phantom{0}0.6/10.0$ & $0.2/\phantom{0}3.3$ & $16.4/72.5$ \\
DFT & $26.3/49.0$ & $64.0/85.8$ & $20.8/41.5$ & $26.9/53.3$ & $\phantom{0}5.4/13.3$ & $3.8/20.0$ & $37.0/80.0$ \\
IDFT & $25.2/48.9$ & $63.1/82.8$ & $20.9/41.9$ & $27.0/53.5$ & $\phantom{0}5.4/23.3$ & $4.2/26.7$ & $30.8/65.0$\\
ASFT & $27.6/54.8$ & $63.8/90.0$ & $21.2/50.4$ & $28.5/59.0$ & $\phantom{0}7.9/23.3$ & $4.2/23.3$ & $39.7/82.5$ \\
PriFT-prob & $30.3/56.2$ & $66.6/\underline{90.8}$ & $22.8/49.6$ & $30.4/59.3$ & $\underline{10.0}/30.0$ & $5.2/20.0$ & $46.6/\underline{87.5}$ \\
PriFT-mass & $30.2/54.1$ & $66.2/90.4$ & $21.7/49.6$ & $\underline{31.4}/57.2$ & $\textbf{10.0}/\underline{36.7}$ & $\underline{5.4}/13.3$ & $46.6/77.5$ \\
\rowcolor{cyan!40!green!10!white} CIG-SFT \rule{0pt}{2.ex} & $\textbf{31.5}/\textbf{58.6}$ & $\textbf{68.4}/89.6$ & $\underline{24.1}/48.5$ & $\textbf{32.1}/59.4$ & $\phantom{0}9.4/\textbf{40.0}$ & $5.2/\underline{26.7}$ & $\textbf{50.0}/\textbf{87.5}$ \\
\midrule
\bottomrule
\end{tabular}
\end{table}




\paragraph{Comparison with reinforcement learning.}
On Qwen2.5-Math-1.5B, CIG-SFT also compares favorably with reinforcement-learning baselines under the same evaluation protocol.
It surpasses GRPO by $3.5/2.4$ and DAPO by $1.0/0.9$, while substantially outperforming the offline preference-optimization baselines DPO and SimPO.
Notably, CIG-SFT obtains these results using only fixed demonstrations, without online rollouts or policy-generated training trajectories.

\paragraph{Code generation.}
Table~\ref{tab:code-main} shows that CIG-SFT's advantage extends beyond mathematical reasoning.
On Qwen2.5-Coder-3B, CIG-SFT achieves the best overall performance among the supervised methods compared, reaching \textbf{$35.4\%$ Avg@16} and \textbf{$50.5\%$ Pass@16}.
This improves over SFT by $2.6/1.5$, DFT by $1.2/1.2$, and the strongest baseline in aggregate performance by $1.1/0.8$.
The gains span HumanEval+, MBPP+, LiveCodeBench, BigCodeBench, and DS-1000, suggesting that the benefit of conditional-information-guided supervision is not specific to mathematical reasoning.


\begin{table}
\centering
\caption{Code-generation results with Qwen2.5-Coder-3B.  We report Avg@16 / Pass@16.} 
\setlength{\tabcolsep}{3.1pt}
\label{tab:code-main}
\begin{tabular}{l|c|cccccc}
\toprule
\midrule
\textbf{Method} & \textbf{Average} & \textbf{HumanEval+} & \textbf{MBPP+} & \textbf{LCB} v5 & \textbf{LCB} v6 & \textbf{BigCode} & \textbf{DS-1000} \\
\midrule
\midrule
Base & $30.9/49.4$ & $40.9/\mathbf{67.1}$ & $49.5/74.1$ & $18.0/28.3$ & $17.1/26.9$ & $35.9/57.2$ & $23.7/42.9$ \\
SFT & $32.8/49.0$ & $44.8/62.2$ & $53.2/\underline{74.3}$ & $18.6/28.2$ & $17.7/26.9$ & $38.2/\underline{59.0}$ & $24.2/43.2$ \\
DFT & $34.2/49.3$ & $43.8/62.2$ & $\underline{54.6}/73.3$ & $21.1/\underline{29.7}$ & $20.0/27.8$ & $\underline{39.9}/58.5$ & $\underline{26.2}/\underline{44.7}$ \\
ASFT  & $\underline{34.3}/\underline{49.7}$ & $45.6/\underline{65.2}$ & $54.2/73.3$ & $\underline{21.1}/29.3$ & $\underline{20.1}/\underline{27.9}$ & $39.5/59.0$ & $25.2/43.7$\\
PriFT-prob & $33.6/49.3$ & $\underline{45.6}/64.0$ & $52.6/73.5$ & $19.9/29.1$ & $19.1/27.8$ & $39.0/56.9$ & $25.6/44.6$ \\
\rowcolor{cyan!40!green!10!white}CIG-SFT \rule{0pt}{2.ex} & $\textbf{35.4}/\textbf{50.5}$ & $\textbf{46.5}/64.0$ & $\textbf{55.0}/\textbf{74.1}$ & $\textbf{22.0}/\textbf{30.5}$ & $\textbf{21.1}/\textbf{29.0}$ & $\textbf{41.0}/\textbf{59.3}$ & $\textbf{26.9}/\textbf{46.4}$\\
\midrule
\bottomrule
\end{tabular}
\end{table}



\begin{table}
\centering
\caption{Ablations on mathematical reasoning with Qwen2.5-Math-1.5B.}
\label{tab:component-ablation}
\setlength{\tabcolsep}{3.2pt}
\begin{tabular}{l|c|cccccc}
\toprule
\midrule
\textbf{Method} & \textbf{Average} & \textbf{MATH500} & \textbf{Minerva} & \textbf{Olympiad} & \textbf{AIME24} & \textbf{AIME25} & \textbf{AMC23} \\
\midrule
\midrule
SFT & $13.8/42.3$ & $41.5/81.8$ & $12.1/40.4$ & $12.0/45.6$ & $\phantom{0}0.6/10.0$ & $0.2/\phantom{0}3.3$ & $16.4/72.5$ \\
\rowcolor{cyan!40!green!10!white} CIG-SFT \rule{0pt}{2.ex} & $31.5/58.6$ & ${68.4}/89.6$ & ${24.1}/48.5$ & ${32.1}/59.4$ & $\phantom{0}9.4/{40.0}$ & $5.2/{26.7}$ & ${50.0}/{87.5}$ \\
\midrule
\rowcolor{cyan!40!blue!10!white}\ \footnotesize{\textcolor{gray}{w/o} $c_t$}& $21.9/48.4$ & $55.4/88.6$ & $19.2/46.0$ & $22.1/54.7$ & $\phantom{0}4.0/16.7$ & $1.4/\phantom{0}6.7$ & $29.4/77.5$\\
\rowcolor{cyan!40!blue!10!white}\ \footnotesize{\textcolor{gray}{w/o} $D_\text{KL}$}& $28.4/55.1$ & $62.1/90.4$ & $19.3/46.0$ & $29.2/59.1$ & $\phantom{0}9.6/30.0$ & $4.4/20.0$ & $45.5/85.0$\\
\midrule
\rowcolor{violet!15!white}\ \footnotesize{Online $c_t$}& $\phantom{0}6.9/21.9$ & $21.1/54.6$ & $\phantom{0}6.6/22.4$ & $\phantom{0}3.3/12.6$ & $\phantom{0}0.2/\phantom{0}3.3$ & $0.2/\phantom{0}3.3$ & $\phantom{0}9.8/24.4$\\
\rowcolor{violet!15!white}\ \footnotesize{Frozen $D_{\text{KL}}$}& $\phantom{}30.5/55.4$ & $64.7/89.4$ & $\phantom{}22.7/47.8$ & $\phantom{}31.2/60.1$ & $11.9/\phantom{}30.0$ & $5.6/\phantom{}20.0$ & $\phantom{}46.7/85.0$\\
\rowcolor{violet!15!white}\ \footnotesize{Both variants}& $\phantom{}15.0/35.1$ & $43.8/76.8$ & $\phantom{}11.8/30.1$ & $\phantom{0}7.6/23.7$ & $\phantom{0}1.9/\phantom{}13.3$ & $0.4/\phantom{0}6.7$ & $\phantom{}24.4/60.0$\\
\midrule
\bottomrule
\end{tabular}
\end{table}
\paragraph{Ablation study.}
Table~\ref{tab:component-ablation} examines both component removal and alternative design choices on Qwen2.5-Math-1.5B. Removing the information ratio (\textcolor{gray}{w/o} $c_t$, eq.~\ref{eq:cigr}) causes the largest degradation among the component ablations, reducing Avg@16$/$Pass@16 from $31.5/58.6$ to $21.9/48.4$, while removing the KL-proximal constraint (\textcolor{gray}{w/o} $D_\text{KL}$, eq.~\ref{eq:projection}) yields a smaller but consistent drop to $28.4/55.1$.
These results indicate that task-conditioned information provides the primary supervision signal, while the proximal projection further improves how this information is incorporated into the current policy.
The design variants further highlight the distinct roles of the frozen reference and current policies.
Computing $c_t$ online with the evolving policy (Online $c_t$) substantially degrades performance to $6.9/21.9$, suggesting that decoupling information estimation from model optimization is important for obtaining a stable attribution signal.
By contrast, replacing the current-policy KL reference with the frozen reference (Frozen $D_{\text{KL}}$) causes only a moderate decline, although it remains inferior to the proposed current-policy projection.
Applying both modifications (Both variants) together reaches only $15.0/35.1$.
Overall, the results support our design of using a frozen policy for offline information attribution and the evolving policy for proximal target construction.


\subsection{Analysis}
\label{sec:exp-sensitivity}

\begin{figure}[t]
    \centering
    \includegraphics[width=0.95\linewidth]{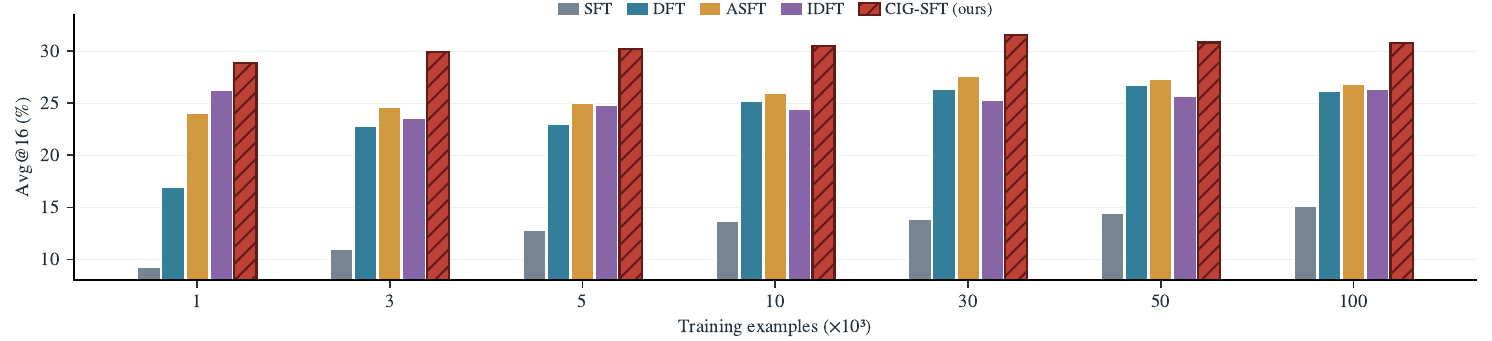}
    \vspace{-4.5mm}
    \caption{Analysis of training scale.}
    \vspace{-3mm}
    \label{fig:datascale}
\end{figure}

\noindent\textbf{Training scale.}\quad
We study the effect of training-set size by varying the number from 1K to 100K.
CIG-SFT consistently achieves the strongest aggregate performance across all data scales, while exhibiting substantially higher data efficiency than the competing objectives.
Notably, with only 1K demonstrations, CIG-SFT already reaches $28.9\%$ Avg@16 and $55.9\%$ Pass@16, outperforming SFT, DFT, ASFT, and IDFT even when they are trained on 100K demonstrations, whose best results are $26.8/53.6\%$.
Moreover, CIG-SFT remains remarkably stable as the training set expands: Avg@16 stays within $[28.9,31.5]\%$ and Pass@16 within $[55.2,58.5]\%$ from 1K to 100K demonstrations.
In contrast, standard SFT and DFT rely more strongly on increasing the amount of training data, while several dynamic SFT baselines exhibit non-monotonic behavior as the dataset grows.
These results suggest that CIG-SFT extracts useful supervision more efficiently from fixed demonstrations and retains its advantage across both low-data and large-scale fine-tuning regimes.

\begin{figure}[t]
    \centering
    \includegraphics[width=0.925\linewidth]{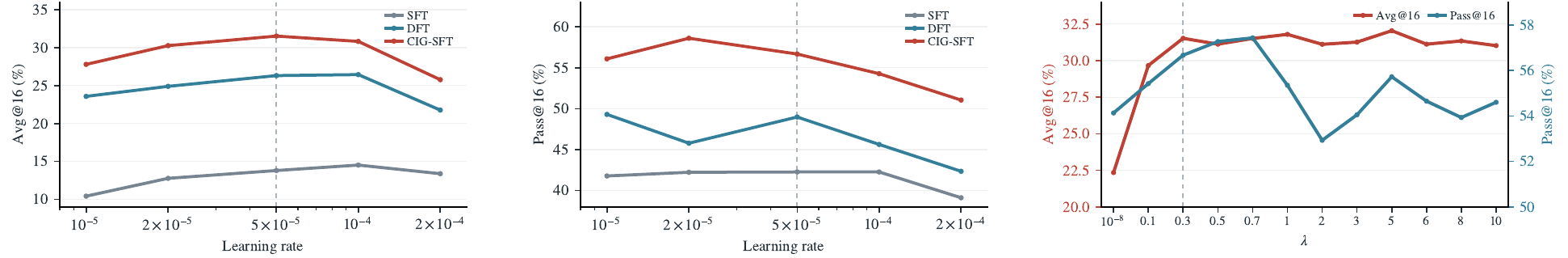}
    \vspace{-4mm}
    \caption{Analysis of learning rate and proximity strength $\lambda$.}
    \vspace{-5.5mm}
    \label{fig:learningrate}
\end{figure}




\begin{figure}[H]
    \centering
    \includegraphics[width=0.945\linewidth]{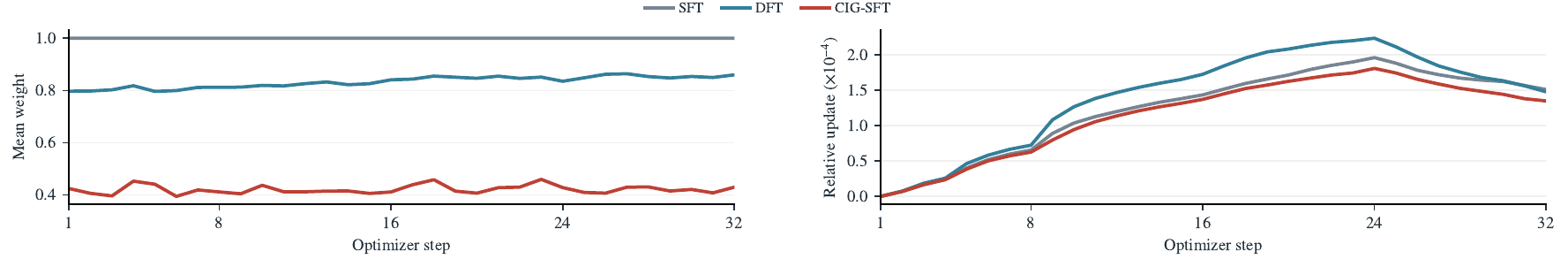}
    \vspace{-4mm}
    \caption{Analysis of the training dynamics.}
    \vspace{-5mm}
    \label{fig:training-dynamics}
\end{figure}

\noindent\textbf{Learning rate.}\quad
We examine sensitivity to the learning rate over
$\{1\times10^{-5}, 2\times10^{-5}, 5\times10^{-5}, 1\times10^{-4}, 2\times10^{-4}\}$.
As shown in Figure~\ref{fig:learningrate}, CIG-SFT consistently outperforms both SFT and DFT across the entire range on Avg@16 and Pass@16.
Its performance remains relatively stable from $2\times10^{-5}$ to $1\times10^{-4}$, with degradation occurring mainly at the largest learning rate.
We use $5\times10^{-5}$ in the main experiments, which provides a favorable balance between Avg@16 and Pass@16.




\noindent\textbf{Proximity strength.}\quad
We vary $\lambda$, which controls how strongly the projected target is constrained to remain close to the current policy.
When $\lambda$ is nearly zero, the proximity constraint effectively vanishes and performance drops substantially, highlighting the importance of controlling the information-induced policy shift.
Once $\lambda$ reaches a moderate value, Avg@16 remains stable over a broad range, with only minor fluctuations from $0.3$ to $10$.
Pass@16 varies more but remains consistently strong in the same region.
We use $\lambda=0.3$ as the default setting in the main experiments.

\noindent\textbf{Training dynamics.}\quad
Figure~\ref{fig:training-dynamics} examines how different objectives allocate supervision and translate it into parameter updates. In the left panel, SFT assigns unit weight to every target token, while DFT maintains a relatively high average weight of approximately
$0.8$-$0.86$. In contrast, CIG-SFT consistently assigns substantially smaller average weights ($\sim0.4$-$0.46$), indicating more selective allocation of learning pressure across demonstrated tokens.
To quantify the resulting optimization magnitude, we report the relative
parameter update at optimizer step $t$,
$
\lVert \Delta\theta_t\rVert
/
\lVert \theta_t\rVert
,
$
where $\Delta\theta_t$ denotes the actual parameter change produced by the $t$-th optimizer step. 
CIG-SFT induces consistently smaller relative parameter updates than DFT and generally smaller updates than standard SFT, while achieving stronger downstream performance. 
This suggests that its gains arise from more selective and better-targeted supervision rather than from more aggressive parameter movement.

\section{Conclusion}
\label{sec:conclusion}

We presented CIG-SFT, an information-theoretic framework that uses task-conditioned information to guide token-level supervision.
It combines conditional information attribution with policy-aware proximal adaptation, yielding a simple gradient-equivalent objective.
Across mathematical reasoning and code generation, CIG-SFT consistently outperforms recent dynamic fine-tuning methods while remaining competitive with reinforcement-learning baselines.
The gains persist across learning rates and data scale, and strong performance can be obtained with substantially fewer demonstrations.
Analysis shows that we assign lower average supervision weights and induce smaller relative parameter updates, suggesting that its benefit comes from more selective allocation of learning pressure rather than more aggressive fitting.
Overall, these results highlight task-conditioned information as a useful complement to model-side compatibility for learning from fixed demonstrations.

\newpage
\section*{AI Use Statement}
Generative AI tools are used as research and writing assistants during the development of this work. Specifically, they are used to assist with checking
and reorganizing mathematical derivations, discussing experimental designs and interpretations, identifying potentially relevant literature, improving the clarity and structure of the manuscript, and assisting with the design of scientific figures.
All AI-assisted technical content is independently reviewed and verified by the authors. The authors check mathematical derivations, manually verify cited literature against the original sources, and validate all reported experimental results and conclusions against the authors' own experiments. The authors take full responsibility for the final content of the paper, including all claims, equations, figures, and experimental results.

\section*{Reproducibility Statement}

We provide the information necessary to reproduce CIG-SFT and its evaluation throughout the paper and supplementary materials.
Section~\ref{sec:method} specifies the complete training objective and its derivation, including the conditional information score, information-proximal projection, and gradient-equivalent objective.
We report the models, datasets, baselines, evaluation metrics, and main hyperparameters used in our experiments.
The training and evaluation code, together with the configuration files required to reproduce the reported experiments, is available in our repository.

\bibliographystyle{iclr2027_conference}
\bibliography{iclr2027_conference}
\clearpage
\appendix

\section{Appendix}
\label{app:appendix}

\subsection{Case Study: Task-Conditioned Credit and Model Probability}
\label{app:case}
\begin{figure}[t]
  \centering
  \includegraphics[width=\linewidth]{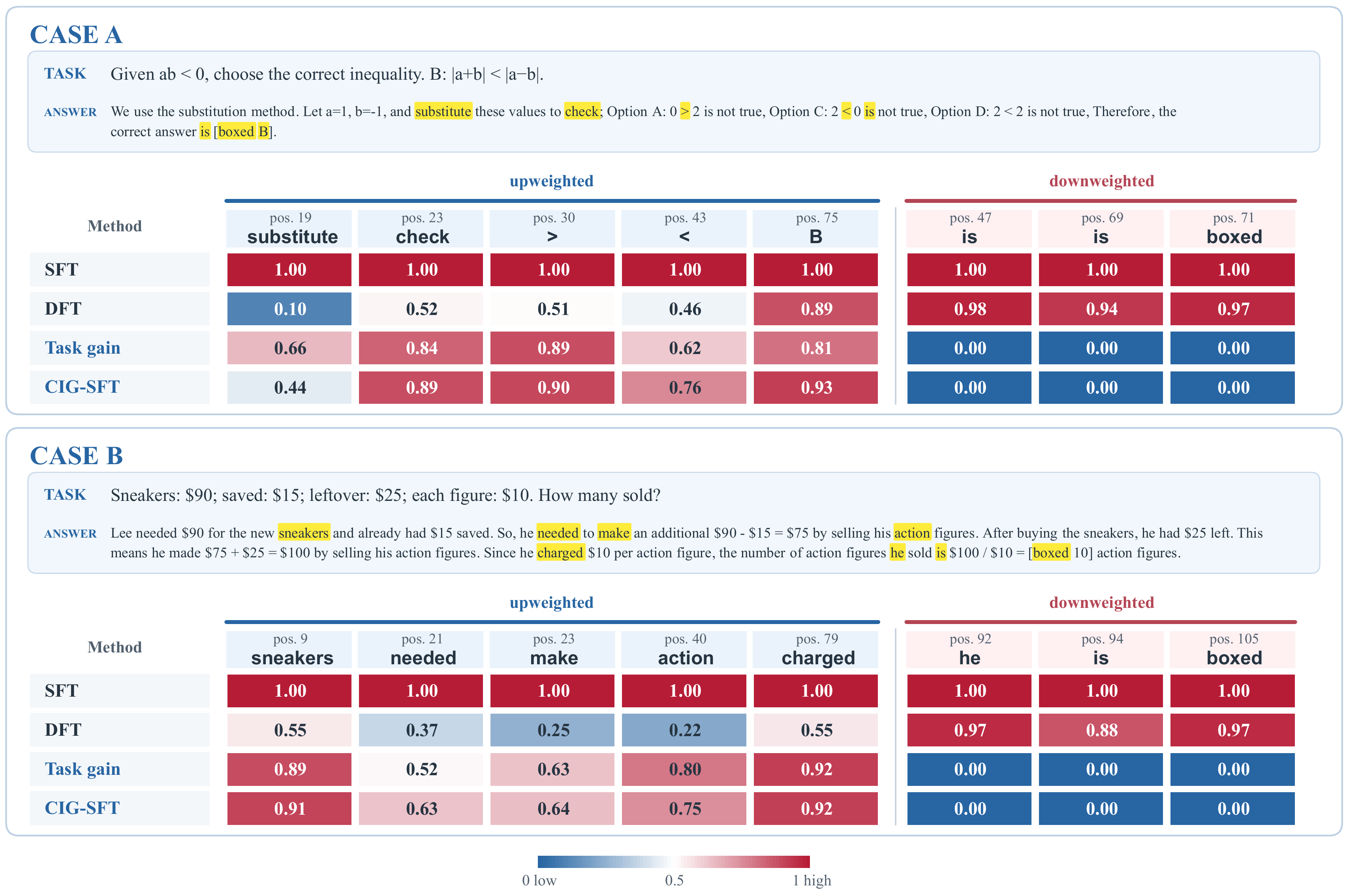}
  \vspace{-5mm}
  \caption{Token-level allocations for two representative cases. The columns show the demonstrated tokens, and the rows report SFT, DFT, task-conditioned information gain $c_t$, and update weights $w_t$.}
  \label{fig:case-a-b-qa}
\end{figure}

For each demonstrated answer token $y_t$, the frozen base model is evaluated under the original task input $x$ and an erased-input counterfactual $\varnothing$, with the gold response prefix held fixed:
\begin{equation}
 c_t=\operatorname{clip}\!\left(\frac{\log \pi_0(y_t\mid x,y_{<t})-\log \pi_0(y_t\mid \varnothing,y_{<t})}{-\log \pi_0(y_t\mid \varnothing,y_{<t})+\varepsilon},0,1\right).
\end{equation}
The resulting $c_t$ values are detached and cached before training. We combine $c_t$ with the current-policy probability to obtain $w_t$. 
Figure~\ref{fig:case-a-b-qa} qualitatively illustrates how CIG-SFT redistributes supervision compared with probability-based weighting.
In Case A, task-relevant tokens such as \textit{substitute}, \textit{check}, the comparison operators, and the final option receive substantial
task-conditioned information gain and are consequently assigned larger weights by CIG-SFT. 
In contrast, highly predictable continuation tokens such as \textit{is} and \textit{boxed} receive near-unit weights under DFT but zero task gain, and are therefore suppressed by CIG-SFT.
Case B exhibits the same pattern: tokens tied to problem entities and arithmetic decisions, such as \textit{sneakers}, \textit{needed}, \textit{action}, and \textit{charged}, are emphasized, whereas generic
continuation and formatting tokens are downweighted.
These examples highlight the distinction between predictive support and task-conditioned information: \textbf{CIG-SFT preferentially allocates supervision to decisions explained by the task input rather than to tokens that are merely easy continuations of the gold prefix}.

\subsection{Computational Cost}
\label{app:cost}

We characterize computational cost using both model-pass accounting and end-to-end measurements. We distinguish trainable forward passes, backward passes, and method-specific preprocessing, since their practical costs can differ substantially.
For a fixed dataset $\mathcal D$, the training cost is
\begin{equation}
T_{\mathrm{train}}(\mathcal D)
=
T_{\mathrm{fwd}}^{\theta}(\mathcal D)
+
T_{\mathrm{bwd}}^{\theta}(\mathcal D)
+
T_{\mathrm{opt}}(\mathcal D)
+
T_{\mathrm{aux}}(\mathcal D).
\end{equation}
Standard SFT requires one trainable forward-backward pass over each training batch. DFT and CIG-SFT reuse the target probabilities from this forward pass and therefore introduce only token-wise scalar operations during training.
CIG-SFT additionally performs one-time offline information scoring with the
frozen reference policy. Specifically, each fixed demonstration is evaluated
once with the original task input and once with an empty task input:
\begin{equation}
T_{\mathrm{preprocess}}
=
T_{\mathrm{fwd}}^{0}(\mathcal D)
+
T_{\mathrm{fwd}}^{0}(\mathcal D^{-}),
\end{equation}
where $\mathcal D^{-}$ denotes the task-free version of the fixed demonstrations.
These passes require no gradient computation or backward pass. The resulting $c_t$ are cached and reused throughout fine-tuning.



We additionally profile the end-to-end cost on Qwen2.5-Math-1.5B using the same 30K-example training set as the main experiments. All methods use a global batch size of 128, micro-batch size of 4, maximum sequence length of 2048, and 233 optimizer steps. Peak GPU memory is measured using \texttt{torch.cuda.max\_memory\_allocated()}.
As shown in Table~\ref{tab:app-measured-cost}, the online training cost of CIG-SFT is essentially identical to standard SFT, with the same peak GPU memory of 28.93 GiB.
The one-time offline scoring stage takes an additional 0.48 hours, giving an end-to-end cost of approximately 2.06 hours when preprocessing is included. 
The cached information scores increase offline storage from 41.5 MiB to 254.2 MiB, an additional 212.7 MiB, while introducing no increase in peak training GPU memory.

\begin{table}[t]
\centering
\caption{Measured training cost.}
\label{tab:app-measured-cost}
\begin{tabular}{lrrr}
\toprule
\textbf{Method} & \textbf{s$/$Step} & \textbf{Peak GiB} & \textbf{Wall clock (h)}\\
\midrule
SFT & $23.92$ & $28.93$ & $1.58$\\
DFT & $23.98$ & $28.93$ & $1.58$\\
\rowcolor{cyan!40!green!10!white} CIG-SFT & $23.92$ & $28.93$ & $0.48+1.58$\\
IDFT & $25.61$ & $42.84$ & $1.68$\\
ASFT & $31.51$ & $40.49$ & $2.07$\\
PriFT-prob & $30.73$ & $31.81$ & $2.03$\\
PriFT-mass & $32.45$ & $31.80$ & $2.14$\\
\bottomrule
\end{tabular}
\end{table}



\subsection{Stability Across Random Seeds}
\label{app:seed}

\begin{table}[t]
\centering
\caption{Stability analysis on Qwen2.5-Math-1.5B.}
\label{tab:app-seed}
\begin{tabular}{lrrrrrrr}
\toprule
\textbf{Method} & \textbf{42} & \textbf{43} & \textbf{44} & \textbf{45} & \textbf{46} & \textbf{Mean $\pm$ Std.} & \textbf{95\% CI}\\
\midrule
SFT & $13.8$ & $13.9$ & $13.9$ & $14.1$ & $14.3$ & $14.01\pm0.20$ & $[13.76,14.26]$\\
DFT & $26.3$ & $26.7$ & $26.2$ & $26.0$ & $26.7$ &  $26.38\pm0.34$ & $[25.95,26.80]$\\
IDFT & $25.2$ & $26.4$ & $26.0$ & $25.6$ & $27.0$ & $26.05\pm0.69$ & $[25.19,26.90]$\\
PriFT-prob & $\underline{30.3}$ & $\underline{29.2}$ & $\underline{29.3}$ & $\underline{29.3}$ & $\underline{29.3}$ & $\underline{29.46}\pm0.46$ & $\underline{[28.89,30.03]}$\\
\rowcolor{cyan!40!green!10!white} CIG-SFT & $\textbf{31.5}$ & $\textbf{30.5}$ & $\textbf{30.6}$ & $\textbf{31.0}$ & $\textbf{31.4}$ & $\textbf{31.00}\pm0.46$ & $[\textbf{30.42},\textbf{31.57}]$\\
\bottomrule
\end{tabular}%
\end{table}

We evaluate optimization stability across five random seeds (42-46) on Qwen2.5-Math-1.5B.
As shown in Table~\ref{tab:app-seed}, CIG-SFT consistently achieves the strongest performance across all seeds, ranging from $30.5\%$ to $31.5\%$ Avg@16.
Its mean performance is $31.00\pm0.46$, with a 95\% confidence interval of $[30.42,31.57]$, indicating limited sensitivity to random initialization and data-order variation.
Notably, the confidence interval remains above that of the strongest competing baseline, PriFT-prob ($[28.89,30.03]$), suggesting that the observed improvement of CIG-SFT is robust rather than driven by a favorable random seed.



\subsection{Detailed Data-Scale Results}
\label{app:scale}

\begin{table}[t]
\centering
\setlength{\tabcolsep}{4.5pt}
\caption{Results of Avg@16 / Pass@16 across training-set sizes.}
\label{tab:app-scale-macro}
\begin{tabular}{lrrrrrrr}
\toprule
\textbf{Method} & \textbf{1K} & \textbf{3K} & \textbf{5K} & \textbf{10K} & \textbf{30K} & \textbf{50K} & \textbf{100K}\\
\midrule
Base & $16.5/51.6$ & $16.5/51.6$ & $16.5/51.6$ & $16.5/\underline{51.6}$ & $16.5/51.6$ & $16.5/51.6$ & $16.5/51.6$\\
SFT & $9.3/38.4$ & $11.0/38.6$ & $12.8/41.2$ & $13.7/43.4$ & $13.8/42.3$ & $14.4/42.5$ & $15.1/42.7$\\
DFT & $16.9/45.5$ & $22.8/46.2$ & $22.9/46.1$ & $25.2/\underline{49.8}$ & $26.3/49.0$ & $26.7/46.9$ & $26.1/47.5$\\
ASFT & $24.0/53.7$ & $\underline{24.6}/\underline{53.1}$ & $\underline{25.0}/\underline{53.4}$ & $\underline{26.0}/51.0$ & $\underline{27.6}/\underline{54.8}$ & $\underline{27.3}/\underline{53.8}$ & $\underline{26.8}/\underline{53.6}$\\
IDFT & $\underline{26.3}/\textbf{57.0}$ & $23.6/47.6$ & $24.8/45.9$ & $24.4/44.4$ & $25.2/48.9$ & $25.7/46.1$ & $26.3/45.3$\\
\rowcolor{cyan!40!green!10!white} CIG-SFT & $\textbf{28.9}/\underline{55.9}$ & $\textbf{29.9}/\textbf{56.4}$ & $\textbf{30.2}/\textbf{58.5}$ & $\textbf{30.5}/\textbf{55.2}$ & $\textbf{31.5}/\textbf{56.7}$ & $\textbf{30.8}/\textbf{56.9}$ & $\textbf{30.8}/\textbf{57.3}$\\
\bottomrule
\end{tabular}
\end{table}

Table~\ref{tab:app-scale-macro} reports the complete numerical results underlying the training-scale analysis in Figure~\ref{fig:datascale}.
We vary the number of training demonstrations from 1K to 100K and report Avg@16/Pass@16 for each method at every scale.
These detailed results complement the aggregate trends shown in the main text and enable direct comparison across individual training-set sizes.


\subsection{Hyperparameter Sensitivity}
\label{app:sensitivity}

\begin{table}[t]
\centering
\caption{Detailed sensitivity results for the learning rate.}
\label{tab:app-lr}
\begin{tabular}{lrrr}
\toprule
\textbf{Learning rate} & \textbf{SFT} & \textbf{DFT} & \textbf{CIG-SFT}\\
\midrule
$1\times10^{-5}$ & $10.4/41.8$ & $23.6/\textbf{49.3}$ & $27.8/56.1$\\
$2\times10^{-5}$ & $12.8/\underline{42.2}$ & $24.9/45.8$ & $30.3/\textbf{58.6}$\\
$5\times10^{-5}$ & $\underline{13.8}/\textbf{42.3}$ & $\underline{26.3}/\underline{49.0}$ & $\textbf{31.5}/\underline{56.7}$\\
$1\times10^{-4}$ & $\textbf{14.5}/\textbf{42.3}$ & $\textbf{26.5}/45.6$ & $\underline{30.8}/54.3$\\
$2\times10^{-4}$ & $13.4/39.1$ & $21.8/42.4$ & $25.8/51.1$\\
\bottomrule
\end{tabular}
\end{table}



\begin{table}[t]
\centering
\setlength{\tabcolsep}{5pt}
\caption{Detailed sensitivity results for the projection coefficient $\lambda$.}
\label{tab:app-lambda}
\begin{tabular}{lrrrrrrrrrrrr}
\toprule
\textbf{$\lambda$} & \textbf{$10^{-8}$} &  \textbf{0.1} & \textbf{0.3} & \textbf{0.5} & \textbf{0.7} & \textbf{1.0} & \textbf{2} & \textbf{3} & \textbf{5} & \textbf{6} & \textbf{8} & \textbf{10} \\
\midrule
Avg@16 & $22.4$ & $29.7$ & $31.5$ & $31.1$ & $31.5$ & $\underline{31.8}$ & $31.1$ & $31.3$ & $\textbf{32.1}$ & $31.1$ & $31.4$ & $31.0$ \\
Pass@16 & $54.1$ & $55.4$ & $56.7$ & $\underline{57.3}$ & $\textbf{57.4}$ & $55.4$ & $52.9$ & $54.1$ & $55.7$ & $54.7$ & $53.9$ & $54.6$ \\
\bottomrule
\end{tabular}%
\end{table}

Tables~\ref{tab:app-lr} and~\ref{tab:app-lambda} provide the complete numerical results underlying the sensitivity analyses shown in Figures~\ref{fig:learningrate} and \ref{fig:training-dynamics}.
Table~\ref{tab:app-lr} reports Avg@16/Pass@16 for SFT, DFT, and CIG-SFT across five learning rates, while Table~\ref{tab:app-lambda} reports the corresponding CIG-SFT results under different values of the projection coefficient $\lambda$.
These tables complement the trends summarized in the main text and provide the exact values used in the sensitivity plots.
Consistent with the main-text analysis, CIG-SFT remains the strongest method across all tested learning rates.
For the projection coefficient, performance drops sharply when $\lambda$ is nearly zero, while remaining relatively stable over a broad range of moderate values.

\subsection{General Capability Retention Analysis}
\label{app:general}

\begin{table}[t]
\centering
\caption{General-task performance. Values are raw Acc (\%).}
\label{tab:app-general-four}
\begin{tabular}{lrrrrr}
\toprule
\textbf{Method} & \textbf{Average} & \textbf{MMLU-General} & \textbf{ARC-Challenge} & \textbf{HellaSwag} & \textbf{PIQA}\\
\midrule
Base & $\textbf{56.1}$ & $\underline{58.2}$ & $\textbf{48.5}$ & $\textbf{45.2}$ & $72.6$ \\
SFT & $\underline{55.9}$ & $57.9$ & $\underline{48.0}$ & $44.6$ & $\textbf{73.0}$ \\
DFT & $54.9$ & $57.9$ & $47.1$ & $42.4$ & $72.0$ \\
IDFT & $54.4$ & $57.7$ & $46.2$ & $41.4$ & $72.2$ \\
ASFT & $55.7$ & $\textbf{58.3}$ & $47.5$ & $44.5$ & $72.6$ \\
\rowcolor{cyan!40!green!10!white} CIG-SFT & $\underline{55.9}$ & $58.0$ & $\underline{48.0}$ & $\underline{44.6}$ & $\underline{72.9}$ \\
\bottomrule
\end{tabular}
\end{table}

We examine whether improving mathematical reasoning comes at the expense of the model's general capabilities. Table~\ref{tab:app-general-four} reports accuracy on four general-purpose benchmarks before and after fine-tuning.
CIG-SFT largely preserves the base model's performance, with the average accuracy changing only from $56.1\%$ to $55.9\%$.
The differences on individual benchmarks are also small: $-0.2$ on MMLU-General, $-0.5$ on ARC-Challenge, $-0.6$ on HellaSwag, and $+0.3$ on PIQA.
In comparison, DFT and IDFT reduce the average accuracy to $54.9\%$ and $54.4\%$, respectively, while ASFT reaches $55.7\%$.
These results provide additional evidence that the reasoning improvements of CIG-SFT are obtained with limited degradation of the general capabilities.




\subsection{In-Domain Generalization}
\label{app:heldout}
\begin{table}[t]
\centering\small
\caption{In domain generalization performance for Qwen2.5-Math-7B.}
\label{tab:app-heldout}
\begin{tabular}{lr}
\toprule
\textbf{Method} & \textbf{Avg@16}\\
\midrule
Base & $32.2$\\
SFT & $42.4$\\
DFT & $\underline{50.8}$\\
IDFT & $49.7$\\
ASFT & $50.5$\\
PriFT-prob & $49.2$\\
PriFT-mass & $47.5$\\
\rowcolor{cyan!40!green!10!white} CIG-SFT & $\textbf{51.8}$\\
\bottomrule
\end{tabular}
\end{table}

We evaluate whether the advantage of CIG-SFT extends to unseen examples drawn from the same distribution as the training data.
Specifically, we randomly sample 1,000 examples from the NuminaMath-CoT data source that are excluded from the training set.
As shown in Table~\ref{tab:app-heldout}, all fine-tuning methods substantially improve over the base model, confirming that they effectively adapt to the training-domain reasoning distribution.
CIG-SFT achieves the best Avg@16 of $51.8\%$, outperforming DFT ($50.8\%$), ASFT ($50.5\%$), and standard SFT ($42.4\%$).
These results indicate that the gains of CIG-SFT are not restricted to out-of-distribution benchmarks: selectively allocating supervision also improves generalization to unseen examples from the training distribution.

\end{document}